\documentclass[runningheads]{llncs}
\usepackage[T1]{fontenc}
\usepackage{hyperref}
\usepackage{amsfonts}
\usepackage{amsmath}
\usepackage{float}
\usepackage{graphicx}
\usepackage{xcolor} 
\usepackage{listings} 
\usepackage{comment}
\usepackage{booktabs}
\usepackage{orcidlink}
\usepackage{multirow}
\usepackage{ulem}

\definecolor{codegreen}{rgb}{0,0.6,0}
\definecolor{codegray}{rgb}{0.5,0.5,0.5}
\definecolor{codepurple}{rgb}{0.58,0,0.82}
\definecolor{backcolour}{rgb}{0.95,0.95,0.92}
\newcommand{\AP}{\mathit{AP}}

\lstdefinestyle{mystyle}{
    backgroundcolor=\color{backcolour},   
    commentstyle=\color{codegreen},
    keywordstyle=\color{magenta},
    numberstyle=\tiny\color{codegray},
    stringstyle=\color{codepurple},
    basicstyle=\ttfamily\footnotesize,
    breakatwhitespace=false,         
    breaklines=true,                 
    captionpos=b,                    
    keepspaces=true,                 
    numbers=left,                    
    numbersep=5pt,                  
    showspaces=false,                
    showstringspaces=false,
    showtabs=false,                  
    tabsize=2
}

\usepackage{tikz}
\usetikzlibrary{shapes, arrows, positioning, shadows, calc}
\usepackage{color}

\begin{document}
\title{NeuroNL2LTL: A Neurosymbolic Framework for Natural Language Translation of Linear Temporal Logic}
\titlerunning{From Logic to Language}
%
\author{Paapa Kwesi Quansah\inst{1,2}\orcidID{\orcidlink{0009-0004-8079-1355}} \and
Ernest Bonnah\inst{1,3}\orcidID{\orcidlink{0000-0001-7170-8936}}}
%
\authorrunning{P. Quansah and E. Bonnah}
%
\institute{Baylor University, Waco TX 76706, USA \and
\email{paapa\_quansah1@baylor.edu} \and
\email{ernest\_bonnah@baylor.edu}}
\maketitle              
\begin{abstract}
Effectively translating between natural language (NL) and formal logics like Linear Temporal Logic (LTL) requires expertise that limits formal verification's reach in safety-critical development. Template-based approaches sacrifice expressiveness for reliability; neural methods achieve fluency but provide no correctness guarantees. We present NeuroNL2LTL, a neurosymbolic architecture unifying learned translation with formal verification. NeuroNL2LTL routes translation through an intermediate representation whose mapping to LTL is structure-preserving by construction. Generated specifications undergo satisfiability and non-triviality checking; a minimal-edit repair mechanism corrects near-miss outputs before they reach downstream tools. The central innovation is verifier-in-the-loop training: verification outcomes serve as reward signals for reinforcement learning, producing neural components that optimize directly for formal correctness. On 200,000+ requirements spanning aerospace, robotics, autonomous vehicles, and ten additional domains, NeuroNL2LTL achieves 28\% semantic equivalence with reference specifications while ensuring 86\% of outputs are verified satisfiable. The system also generates contextually grounded explanations from LTL, enabling domain experts to validate specifications without specialized training. This work demonstrates that formal verification can function as both training objective and runtime filter for neural specification systems, allowing us to build neural-based tools whose reliability derives from logical guarantees rather than statistical confidence.

\end{abstract}

\section{Introduction}
\label{sec:introduction}


Temporal formalisms such as Linear Temporal Logic (LTL) \cite{pnueli1977temporal} provide a rich semantics to unambiguously specify liveness and safety properties, sequencing constraints, response guarantees in safety-critical hardware, software, and communication systems \cite{rozier2011linear},  dynamical systems~\cite{cai2021modular,xie2020temporal}, etc. Using model checkers like SPIN \cite{holzmann2004spin}, Spot \cite{duret2016spot}, etc., and theorem provers, we can then verify system designs against these specifications with mathematical guarantee certainty and correctness \cite{clarke2018handbook}. Despite the expressive power of LTL, formalizing requirements as LTL specifications requires specialized mathematical expertise and familiarity. However, while domain experts who understand system behavior rarely possess formal methods expertise, formal methods specialists rarely possess domain knowledge of system operation. This mismatch introduces requirement translation and specification errors in succinctly and compactly formalizing behavior-critical systems.

This paper introduces NeuroNL2LTL, a neurosymbolic framework for contextually grounded translation of requirements between NL and LTL. Our proposed tool accepts complex, unrestricted requirements alongside domain context: definitions of atomic propositions, domain labels, and semantic grounding information. The architecture decomposes translation into two stages. First, a neural encoder maps NL requirements to an Intermediate Technical Language (ITL) whose syntax mirrors LTL's logical structure while remaining human-readable. A deterministic converter then produces LTL from ITL through structure-preserving transformation. This decomposition localizes potential errors: the neural component may produce malformed or incorrect ITL, but the ITL-to-LTL mapping is sound by construction. At inference time, the Spot model checker \cite{duret2016spot} verifies each generated formula for satisfiability and non-triviality, filtering outputs that would fail basic sanity checks. A minimal-edit repair module attempts to correct near-miss generations before rejection. One of the system's central methodological contributions is verifier-in-the-loop training. Rather than optimizing neural parameters against reference translations alone, our tool incorporates verification outcomes as reinforcement learning rewards. The neural encoder learns to produce ITL that survives formal checking, not merely ITL that resembles training examples.

We train our proposed framework on a corpus of over 200,000 requirement-specification pairs spanning thirteen domains including aerospace, autonomous vehicles, robotics, and medical devices. Each training instance includes domain context: definitions mapping atomic propositions to their domain-specific meanings. The system achieves 28\% semantic equivalence with reference LTL specifications on held-out data, with 86\% of generated formulas verified satisfiable and non-trivial. Ablation experiments isolate the contribution of each architectural component: removing verifier-in-the-loop training reduces semantic equivalence over 30 relative percentage points; disabling repair reduces syntactic correctness by 10 points. NeuroNL2LTL also generates natural language explanations from LTL, grounding explanations in domain-specific proposition definitions so that domain experts can validate specifications without reading temporal logic directly. Human evaluation confirms that generated explanations are fluent, accurate, and contextually appropriate.

The contributions of this work are as follows. We present the first neurosymbolic architecture for contextually grounded translation that uses formal verification outcomes to directly optimize neural translation toward logical correctness. We introduce an intermediate representation that isolates neural uncertainty from deterministic logical transformation, enabling targeted repair when generation fails. We demonstrate that satisfiability and non-triviality checking function as effective runtime filters, catching classes of errors that supervised training alone does not prevent. We provide a methodology for contextual LTL explanation generation, validated through human evaluation across technical domains. Together, these contributions establish that formal verification can serve as supervision signal for learning systems, a principle applicable beyond LTL to any domain where neural outputs admit formal checking.

\section{Preliminaries}
\label{sec:preliminaries}


Let $\AP$ be a finite set of \textit{atomic propositions} and $\Sigma = 2^{\AP}$ be the powerset of $\AP$. For any alphabet $A$, we write $A^*$ for the set of finite sequences over $A$ and $A^\omega$ for the set of infinite sequences. A \textit{trace}, $t \in \Sigma^\omega$, is an infinite sequence of events $t = e_0, e_1, e_2, \dots$, where $e_{i} \in \Sigma$ and $i \in \mathbb{N}$. For a trace $t$, we represent the $i^{th}$ event by $t[i]$. 

\vspace{-2mm}
\subsection{Linear Temporal Logic}
Linear Temporal Logic (LTL) extends propositional logic with temporal operators that express properties over infinite sequences of states \cite{pnueli1977temporal}. The syntax of LTL formulas is defined inductively as:
\[
\varphi \mathrel{::=} \top \mid p \mid \neg \varphi \mid \varphi_{1} \land \varphi_{2} \mid \mathbf{X}\varphi \mid \mathbf{G}\varphi \mid \mathbf{F}\varphi \mid \varphi_{1} \mathbin{\mathbf{U}} \varphi_{2} 
\]

\noindent where $\top$ stands for true, $p$ is an atomic proposition in $AP$. The operators $\neg$, and $\land$ represent the conjunction and negation Boolean operators respectively. The operators $\mathbf{X}$, $\mathbf{G}$, $\mathbf{F}$, and $\mathbf{U}$ denote the next, globally, eventually, and until operators respectively. The disjunction operator ($\vee$) can be derived from the negation and conjunction operators. Likewise, the implication operator ($\rightarrow$) can also be derived from the negation and disjunction operators. Similarly, non-primitive operators such as $\mathbf{W}$ (weak until) and $\mathbf{R}$ (release) can be derived from the standard operators defined in the grammar.

\vspace{-2mm}
\subsection{Verification Properties}
The symbolic backend of the proposed framework verifies two properties of the generated formulas. A formula $\varphi$ is \textit{satisfiable} if the language defined by $\mathcal{L}(\varphi) = \{t \mid t[0] \models \varphi\}$ is non-empty. A formula is \textit{non-trivial} if it is neither valid nor unsatisfiable; formally, $\mathcal{L}(\varphi) \neq \emptyset$ and $\mathcal{L}(\varphi) \neq (2^{AP})^\omega$. Trivial formulas, comprising tautologies and contradictions, typically signal translation failures because requirements that hold vacuously or preclude all executions cannot capture genuine system constraints.

We then employ the Spot library~\cite{duret2004spot} for verification. Specifically, Spot translates LTL formulas into Büchi automata to evaluate language emptiness. To establish semantic equivalence, our framework verifies that $\mathcal{L}(\varphi_{1}) = \mathcal{L}(\varphi_{2})$ by confirming the emptiness of the symmetric difference: $\mathcal{L}(\varphi_{1} \land \neg\varphi_{2}) = \emptyset$ and $\mathcal{L}(\neg\varphi_{1} \land \varphi_{2}) = \emptyset$.

\subsection{The Contextual Grounding Problem}
The translation problem addressed by our proposed framework is not syntax-to-syntax conversion but contextually grounded translation. A NL requirement $r$ is accompanied by a domain context $\mathcal{C}$ that defines the interpretation of atomic propositions. Formally, $\mathcal{C}: AP \rightarrow \textsc{Descriptions}$ maps each proposition to a natural language description of the condition it represents. For example, in an automotive domain: $\mathcal{C}(p) = \text{``lane departure detected''};  \mathcal{C}(q) = \text{``obstacle detection}$ \\$\text{active''}; \mathcal{C}(r) = \text{``driver override requested''}$

Given requirement $r$ and context $\mathcal{C}$, the translation task is to generate an LTL formula $\varphi$ over $AP$ such that $\varphi$ captures the temporal constraints expressed in $r$ with propositions interpreted according to $\mathcal{C}$. The reverse task produces a natural language explanation $r'$ from $\varphi$ that grounds temporal structure in the domain-specific meanings provided by $\mathcal{C}$. This formulation distinguishes contextually grounded translation from context-free approaches that treat propositions as abstract symbols. A system without access to $\mathcal{C}$ cannot determine whether ``the sensor must eventually calibrate after detecting an obstacle'' should translate to $\mathbf{G}(q \rightarrow \mathbf{F} s)$ or $\mathbf{G}(p \rightarrow \mathbf{F} r)$; the mapping depends on which proposition denotes sensor calibration in the target domain.

\subsection{Intermediate Technical Language}
Our framework routes translation through an Intermediate Technical Language (ITL) that mirrors LTL's logical structure in human-readable form. The ITL grammar is defined by a deterministic mapping $\mathcal{T}: \textsc{LTL} \rightarrow \textsc{ITL}$ that operates recursively on the abstract syntax tree of an LTL formula:
\begin{align*}
\mathcal{T}(p) &= \text{``}p\text{''} & \mathcal{T}(\top) &= \text{``true''} & \mathcal{T}(\bot) &= \text{``false''} \\
\mathcal{T}(\neg\varphi) &= \text{``not ''} \cdot \mathcal{T}(\varphi) \\
\mathcal{T}(\varphi_{1} \land \varphi_{2}) &= \mathcal{T}(\varphi_{1}) \cdot \text{`` and ''} \cdot \mathcal{T}(\varphi_{2}) \\
\mathcal{T}(\varphi_{1} \lor \varphi_{2}) &= \mathcal{T}(\varphi_{1}) \cdot \text{`` or ''} \cdot \mathcal{T}(\varphi_{2}) \\
\mathcal{T}(\varphi_{1} \rightarrow \varphi_{2}) &= \text{``if ''} \cdot \mathcal{T}(\varphi_{1}) \cdot \text{``, then ''} \cdot \mathcal{T}(\varphi_{2}) \\
\mathcal{T}(\varphi_{1} \leftrightarrow \varphi_{2}) &= \mathcal{T}(\varphi_{1}) \cdot \text{`` if and only if ''} \cdot \mathcal{T}(\varphi_{2}) \\
\mathcal{T}(\mathbf{X}\varphi) &= \text{``in the next state, ''} \cdot \mathcal{T}(\varphi) \\
\mathcal{T}(\mathbf{G}\varphi) &= \text{``always, ''} \cdot \mathcal{T}(\varphi) \\
\mathcal{T}(\mathbf{F}\varphi) &= \text{``eventually, ''} \cdot \mathcal{T}(\varphi) \\
\mathcal{T}(\varphi_{1} \mathbin{\mathbf{U}} \varphi_{2}) &= \mathcal{T}(\varphi_{1}) \cdot \text{`` until ''} \cdot \mathcal{T}(\varphi_{2}) \\
\mathcal{T}(\varphi_{1} \mathbin{\mathbf{R}} \varphi_{2}) &= \mathcal{T}(\varphi_{1}) \cdot \text{`` releases ''} \cdot \mathcal{T}(\varphi_{2}) \\
\mathcal{T}(\varphi_{1} \mathbin{\mathbf{W}} \varphi_{2}) &= \mathcal{T}(\varphi_{1}) \cdot \text{`` weakly until ''} \cdot \mathcal{T}(\varphi_{2})
\end{align*}
\noindent where $\cdot$ denotes string concatenation. The mapping $\mathcal{T}$ is total over the LTL fragment and invertible: an ITL string can be parsed back to LTL by reversing the keyword mappings and reconstructing the AST. We denote the inverse $\mathcal{T}^{-1}: \textsc{ITL} \rightarrow \textsc{LTL}$.

\noindent \textbf{Proposition 1}(\textit{Structure Preservation})
For any LTL formula $\varphi$: $\mathcal{T}^{-1}(\mathcal{T}(\varphi)) \equiv \varphi$.

\textit{Proof sketch:} The proof follows by structural induction on $\varphi$. Each LTL operator maps to a unique ITL keyword, and the recursive structure of $\mathcal{T}$ preserves operator scope and precedence. The round-trip property ensures that ITL functions as a lossless intermediate representation. ITL serves two purposes in architecture. First, it provides a more natural target for neural generation than raw LTL syntax: keywords like ``until'' and ``eventually'' are closer to natural language than symbols like $\mathbf{U}$ and $\mathbf{F}$, reducing the representational distance the neural encoder must bridge. Second, it localizes the source of translation errors. If the neural component produces malformed output, the error lies in the NL-to-ITL mapping; the ITL-to-LTL conversion is deterministic and correct by construction. This decomposition enables targeted repair: near-miss ITL strings can be corrected without reasoning about LTL syntax directly.


\section{The System Architecture}
\label{sec:architecture}
\subsection{System Overview}
\label{subsec:overview}
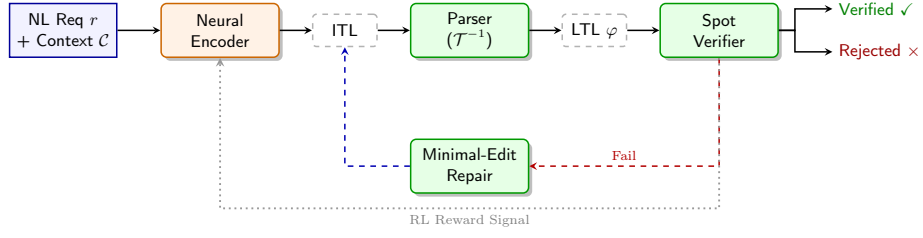
\begin{figure}[t]
\centering
\resizebox{\textwidth}{!}{%
\begin{tikzpicture}[
    node distance=1.2cm and 1.0cm, 
    font=\small\sffamily,
    >=stealth,
    thick,
    neural/.style={
        rectangle, 
        rounded corners=3pt, 
        draw=orange!80!black, 
        fill=orange!10, 
        minimum height=1cm, 
        minimum width=2.2cm, 
        align=center,
        drop shadow
    },
    symbolic/.style={
        rectangle, 
        rounded corners=3pt, 
        draw=green!60!black, 
        fill=green!10, 
        minimum height=1cm, 
        minimum width=2.2cm, 
        align=center,
        drop shadow
    },
    data/.style={
        rectangle, 
        rounded corners=2pt, 
        draw=gray!60, 
        fill=white, 
        minimum height=0.6cm, 
        minimum width=1.2cm, 
        align=center,
        dashed
    },
    input/.style={
        rectangle, 
        draw=blue!60!black, 
        fill=blue!5, 
        minimum height=1cm, 
        minimum width=2cm, 
        align=center
    }
]

    \node[input] (input) {NL Req $r$ \\ + Context $\mathcal{C}$};
    \node[neural, right=0.8cm of input] (encoder) {Neural\\Encoder};
    \node[data, right=0.6cm of encoder] (itl) {ITL};
    \node[symbolic, right=0.6cm of itl] (parser) {Parser\\($\mathcal{T}^{-1}$)};
    \node[data, right=0.6cm of parser] (ltl) {LTL $\varphi$};
    \node[symbolic, right=0.6cm of ltl] (spot) {Spot\\Verifier};

    \node[right=1cm of spot, yshift=0.4cm, text=green!50!black, anchor=west] (valid) {Verified $\checkmark$};
    \node[right=1cm of spot, yshift=-0.4cm, text=red!60!black, anchor=west] (invalid) {Rejected $\times$};
    \node[symbolic, below=1.5cm of parser] (repair) {Minimal-Edit\\Repair};

    \draw[->] (input) -- (encoder);
    \draw[->] (encoder) -- (itl);
    \draw[->] (itl) -- (parser);
    \draw[->] (parser) -- (ltl);
    \draw[->] (ltl) -- (spot);

    \draw[->] (spot.east) -- ++(0.3,0) |- (valid);
    \draw[->] (spot.east) -- ++(0.3,0) |- (invalid);

    \draw[->, dashed, draw=red!70!black] (spot.south) 
        |- node[pos=0.75, above, font=\scriptsize, text=red!70!black] {Fail} (repair.east);
    
    \draw[->, dashed, draw=blue!70!black] (repair.west) 
        -| (itl.south);

    \draw[->, dotted, draw=gray!80, line width=1pt] (spot.south) 
        -- ++(0,-2.8) -| node[pos=0.25, below, font=\scriptsize, text=gray] {RL Reward Signal} (encoder.south);

\end{tikzpicture}
}
\caption{The architecture of NeuroNL2LTL framework} 
\label{fig:architecture}
\vspace{-6mm}
\end{figure}

NeuroNL2LTL translates between natural language requirements and Linear Temporal Logic through a neurosymbolic architecture that decomposes the problem into neural and symbolic components with clearly defined interfaces. The neural components handle the ambiguity inherent in natural language while the symbolic components enforce logical correctness. This section presents the architecture, details each component, and explains how formal verification integrates into the training process. As shown in Figure~\ref{fig:architecture}, our proposed neurosymbolic framework comprises four stages, namely, neural encoding, ITL-to-LTL conversion, LTL verification, and minimal-edit repair. The details of the stages are discussed below. 




\noindent \uline{\textit{Neural Encoding:}}
The neural encoder implements a sequence-to-sequence model that maps natural language to ITL. The encoder receives a structured input consisting of three elements: the natural language requirement $r$, a domain label $d$ (e.g., ``aerospace'', ``robotics''), and the context mapping $\mathcal{C}$ rendered as natural language definitions of atomic propositions. We initialize the encoder from a pretrained large language model (Flan-T5-XL) that has undergone continued pretraining on a corpus of approximately five billion tokens drawn from technical documentation across the thirteen target domains. This domain adaptation improves the model's handling of specialized terminology and temporal phrasing patterns common in requirements documents. The adapted model is then fine-tuned on more than $200k+$ paired (NL, ITL) examples using standard supervised learning before the verifier-in-the-loop training stage described in Section~\ref{subsec:vil}. At inference time, we apply grammar-constrained decoding to bias generation toward syntactically valid ITL. The decoder maintains a set of valid next tokens at each generation step according to the ITL grammar, assigning zero probability to tokens that would produce unparseable output. This constraint reduces but does not eliminate syntactic errors; the repair module handles cases where constrained decoding fails. The neural decoder for LTL-to-NL translation shares the same base architecture but is initialized in a different mode. It receives an ITL string (obtained by applying $\mathcal{T}$ to the input LTL) together with domain context $\mathcal{C}$ and generates natural language. Conditioning on both logical structure and domain definitions enables the decoder to produce explanations grounded in application-specific terminology rather than abstract logical paraphrase. 

\noindent \uline{\textit{Deterministic ITL-to-LTL Conversion:}}
The parser implements the inverse mapping $\mathcal{T}^{-1}$ from ITL strings to LTL formulas. We implement the parser using Lark, a parsing toolkit that generates parsers from EBNF grammars. The ITL grammar is unambiguous: each valid ITL string corresponds to exactly one parse tree, and the parse tree determines a unique LTL formula. Parsing proceeds in two phases, i.e., lexical analysis and syntactic analysis. The lexical analysis tokenizes the input string, identifying ITL keywords (``always'', ``eventually'', ``until'', ``releases'', ``weakly until'', ``if'', ``then'', ``and'', ``or'', ``not'', ``in the next state'', ``if and only if'') and atomic proposition identifiers. Syntactic analysis constructs an abstract syntax tree according to the grammar. The AST is then traversed to emit the corresponding LTL formula string in Spot's input format.

\noindent \textbf{Proposition 2} (\textit{Parser Correctness})
For any ITL string $s$ in the language of the ITL grammar, if the parser produces LTL formula $\varphi$, then $\mathcal{T}(\varphi) = s$.

\textit{Proof Sketch} This property holds by construction: the parser inverts each production rule of $\mathcal{T}$. We verified the property empirically by round-trip testing on the full dataset: for each LTL formula $\varphi$, we computed $\mathcal{T}(\varphi)$, parsed the result to obtain $\varphi'$, and confirmed $\varphi \equiv \varphi'$ using Spot's equivalence checker. All 15,000,000 unique formulas generated at this stage passed this test. If parsing fails, the input string violates the ITL grammar. The parser returns a structured error indicating the failure location and expected tokens, which guides the repair module. 

\noindent \uline{\textit{Verification Backend}}
Successfully parsed LTL formulas undergo verification using the Spot library. Spot converts LTL formulas to equivalent Büchi automata and provides decision procedures for standard properties. NeuroNL2LTL checks two properties:

 \textit{Satisfiability.} A formula $\varphi$ is satisfiable if $\mathcal{L}(\varphi) \neq \emptyset$. Spot determines satisfiability by constructing the Büchi automaton for $\varphi$ and checking whether the automaton accepts any infinite word. An unsatisfiable formula represents a contradictory requirement that no system execution can satisfy.

\textit{Non-triviality.} A formula $\varphi$ is non-trivial if it is neither unsatisfiable nor valid. We check validity by testing whether $\neg\varphi$ is unsatisfiable. A valid formula (tautology) provides no constraint on system behavior; a translation that produces $\top$ or a logical equivalent has failed to capture the requirement's content.

Formulas that pass both checks are emitted as verified output. Formulas that fail either check are passed to the repair module with an error classification (unsatisfiable, trivial, or syntactically malformed from an earlier parsing failure). The verification backend also supports semantic equivalence checking for evaluation. Given formulas $\varphi_{1}$ and $\varphi_{2}$, Spot tests whether $\mathcal{L}(\varphi_{1}) = \mathcal{L}(\varphi_{2})$ by checking that both $\varphi_{1} \land \neg\varphi_{2}$ and $\neg\varphi_{1} \land \varphi_{2}$ are unsatisfiable. This equivalence check is the primary metric for translation accuracy: a generated formula is correct if it is semantically equivalent to the reference formula.

\noindent \uline{\textit{Minimal-Edit Repair}}
\label{subsec:repair}
When parsing or verification fails, the repair module attempts to correct the error through minimal edits. The module operates in two layers: a fast heuristic layer for common errors and a learned layer for structural errors that heuristics cannot resolve.

\textit{Heuristic repair.} The first layer addresses syntactic errors that arise from predictable failure modes: unbalanced parentheses, missing operators between subformulas, and operator precedence ambiguities. The heuristics apply a ranked sequence of edits: parenthesis insertion or deletion at the error location, operator insertion between adjacent propositions, and keyword normalization (e.g., correcting ``eventual'' to ``eventually''). Each edit is validated by re-parsing; the first edit that yields a parseable ITL string is accepted. If no single edit succeeds within a fixed budget of $m$ attempts, control passes to the learned layer.

\textit{Learned repair.} The second layer employs a graph neural network trained to correct malformed ITL abstract syntax trees. When an ITL string parses partially (producing a partial AST with error nodes), the GNN receives the partial AST as input and predicts a sequence of edit operations: node relabeling, edge redirection, subtree deletion, or subtree insertion. The GNN was trained on pairs of malformed and corrected ASTs collected during NeuroNL2LTL's development, with successful repairs (those producing verifiable LTL) providing positive training signal.

The repair module returns either a corrected ITL string or a failure indication. Repaired strings re-enter the pipeline at the parsing stage. If repair fails after both layers, the translation is rejected, and no output is returned. Repair cost, defined as the number of edit operations applied, provides a training signal: formulas requiring extensive repair indicate unreliable neural generation, while formulas requiring no repair indicate successful translation. The verifier-in-the-loop training stage (described in Section~\ref{subsec:vil}) penalizes high repair costs to incentivize direct generation of correct formulas.
\vspace{-3mm}

\subsection{Verifier-in-the-Loop Training}
\label{subsec:vil}
\vspace{-2mm}
Supervised training on paired (NL, ITL) examples teaches the neural encoder to produce outputs that resemble training data. This objective is necessary but insufficient: a model can achieve low loss on reference examples while still generating formulas that fail verification. Verifier-in-the-loop training addresses this gap by incorporating verification outcomes directly into the learning objective.

We formulate the problem as reinforcement learning. The neural encoder acts as a policy $\pi_\theta$ that maps input (requirement, context) pairs to ITL strings. The symbolic backend provides a reward signal $R$ based on the generated ITL and its corresponding LTL. The training objective maximizes expected reward:
$\mathcal{J}(\theta) = \mathbb{E}_{(r, \mathcal{C}) \sim \mathcal{D}} \left[ \mathbb{E}_{s \sim \pi_\theta(\cdot \mid r, \mathcal{C})} \left[ R(s) \right] \right]$,
\noindent where $\mathcal{D}$ is the training distribution over requirements and contexts.

The reward function combines three components derived from the symbolic backend:
\begin{align*}
R(s) &= \alpha \cdot \mathbb{1}[\text{parse}(s) \text{ succeeds}] \\
&\quad + \beta \cdot \mathbb{1}[\text{sat}(\mathcal{T}^{-1}(s)) \land \text{nontriv}(\mathcal{T}^{-1}(s))] \\
&\quad - \gamma \cdot \text{repair\_cost}(s)
\end{align*}

\noindent where $\alpha$, $\beta$, $\gamma$ are weighting coefficients. The first term rewards syntactically valid ITL. The second term rewards LTL formulas that pass satisfiability and non-triviality checks. The third term penalizes formulas that required repair, incentivizing direct generation of correct output. We optimize $\mathcal{J}(\theta)$ using Group Relative Policy Optimization (GRPO), a variant of policy gradient methods that computes advantages relative to a group of samples from the same input. For each training example, we sample multiple ITL candidates from $\pi_\theta$, compute rewards for each, and update parameters to increase the probability of higher-reward candidates relative to lower-reward ones.

This training procedure directly optimizes for formal properties checked by the symbolic backend. The neural encoder then learns to produce strings that survive parsing, convert to satisfiable non-trivial LTL, and require minimal repair rather than merely to imitate reference ITL strings. Ablation experiments (Section~\ref{sec:evaluation}) demonstrate that verifier-in-the-loop training improves semantic equivalence by over 30 relative percentage points compared to supervised training alone.
\vspace{-2mm}

\subsection{Correctness Considerations}
\label{subsec:correctness}
NeuroNL2LTL provides guarantees at different levels of the architecture. The ITL-to-LTL conversion is correct by construction: the parser implements $\mathcal{T}^{-1}$ exactly, and round-trip testing confirms that $\mathcal{T}^{-1}(\mathcal{T}(\varphi)) \equiv \varphi$ for all formulas in the dataset. The verification backend inherits Spot's correctness: satisfiability and non-triviality checks are sound and complete for LTL over infinite traces.

The neural encoder provides no formal guarantees. A generated ITL string that passes verification is guaranteed to be satisfiable and non-trivial, but it is not guaranteed to capture the intended meaning of the input requirement. Semantic equivalence with the intended specification depends on the accuracy of neural translation, which our evaluation measures empirically.

This separation clarifies the role of each component. The symbolic backend acts as a filter: it cannot verify that a translation is correct (in the sense of matching intent), but it can verify that a translation is not obviously wrong (in the sense of being unsatisfiable or trivial). Verifier-in-the-loop training improves the neural encoder's reliability, but the final output remains a proposal that downstream processes, including human review, may validate further.
\vspace{-3mm}

\section{Evaluation}
\label{sec:evaluation}
\vspace{-3mm}
This section presents an empirical evaluation of NeuroNL2LTL across multiple dimensions: translation accuracy, comparison with baseline systems, effectiveness of the verification filter, repair module performance, and ablation studies isolating individual architectural contributions.

\subsection{Experimental Setup}
\label{subsec:setup}

\paragraph{Dataset.} We evaluate on the VERIFY corpus comprising 218,871 requirement-specification pairs across thirteen domains: aerospace, automotive/autonomous vehicles, robotics, medical devices, industrial automation, home automation, smart grid/energy management, financial/transaction systems, networking/distributed systems, security/authentication, build pipelines/CI-CD, version control, and web services/APIs. Each instance contains a natural language requirement, domain context (natural language definitions for atomic propositions), the canonical ITL representation, and the reference LTL formula verified by Spot. The dataset spans LTL formulas with abstract syntax tree depths from 1 to 22, with a mean depth of 5.0. We partition the data into complexity strata: simple (depth 1--4, 31\% of instances), medium (depth 5--8, 42\%), high (depth 9--12, 19\%), and very high (depth 13+, 8\%). We reserve 10\% of each stratum for testing, ensuring the test set reflects the full complexity distribution.

\noindent\textbf{Metrics.} We evaluate using four metrics:

\textit{Semantic Equivalence (SemEq)}: The primary metric. A generated formula $\varphi_g$ is semantically equivalent to the reference $\varphi_r$ if $\mathcal{L}(\varphi_g) = \mathcal{L}(\varphi_r)$, verified using Spot's equivalence checker. We report the percentage of test instances where the generated formula is semantically equivalent to the reference.

\textit{Syntactic Correctness (SynCorr)}: The percentage of outputs that parse successfully as valid ITL and convert to well-formed LTL. This measures whether the neural encoder produces structurally valid output before considering semantic correctness.

\textit{Satisfiability (Sat)}: The percentage of syntactically correct outputs that Spot verifies as satisfiable (non-contradictory).

\textit{Non-triviality (NonTriv)}: The percentage of satisfiable outputs that are also non-trivial (neither tautologies nor contradictions).

Note that Sat and NonTriv are conditional metrics: Sat conditions on syntactic correctness, and NonTriv conditions on satisfiability. The unconditional rate of outputs passing all three checks (syntactically correct, satisfiable, and non-trivial) is the product SynCorr $\times$ Sat $\times$ NonTriv.

\noindent \textbf{Baselines.} We compare NeuroNL2LTL against two categories of baselines.

\textit{Large Language Models (zero-shot and few-shot):} We evaluate GPT-4o, Gemini 1.5 Pro, and Claude 3.5 Sonnet in both zero-shot and 5-shot configurations. For zero-shot, we provide the task description, domain context, and requirement, prompting the model to output LTL directly. For 5-shot, we prepend five (requirement, context, LTL) examples from the training set, selected to cover diverse temporal operators. We use greedy decoding (temperature 0) for reproducibility. We compare NeuroNL2LTL against four systems from the literature: Lang2LTL \cite{pan2023data}, NL2TL \cite{chen2023nl2tl}, CopyNet-LTL \cite{liu2023grounding}, and a Seq2Seq-Attn baseline trained on VERIFY. Note that Lang2LTL, NL2TL, and CopyNet-LTL were designed for simpler, domain-specific commands (primarily robot navigation) without contextual grounding. Our evaluation on complex, multi-domain requirements with explicit proposition definitions represents a distribution shift for these systems.

\textit{Implementation.} NeuroNL2LTL uses Flan-T5-XL (3B parameters) as the base model, with continued pretraining (CPT) on 5B tokens of technical documentation followed by supervised fine-tuning on 180k training pairs and verifier-in-the-loop training using GRPO for 10k steps. We use the Spot library (version 2.11.6) for all verification. The repair module's GNN uses a 4-layer graph attention network with 256-dimensional hidden states. All experiments run on 4$\times$ NVIDIA H200 GPUs.

\vspace{-3mm}
\subsection{Main Results: Translation Accuracy}
\label{subsec:main_results}
\vspace{-2mm}
We present the results on the performance of NeuroNL2LTL on the test set. Table~\ref{tab:main_results} reports syntactic correctness, satisfiability, and non-triviality stratified by formula complexity, and Table~\ref{tab:baseline_comparison} reports semantic equivalence alongside baselines. NeuroNL2LTL achieves 27.8\% semantic equivalence overall, with 93.7\% of outputs syntactically correct. Of syntactically correct outputs, 96.8\% are satisfiable and 95.4\% of those are non-trivial, yielding an unconditional verification pass rate (syntactically correct, satisfiable, and non-trivial) of 86.2\%. Syntactic correctness remains high (87.6\%+) even at very high complexity, indicating that the grammar-constrained decoding and repair mechanisms successfully produce parseable output regardless of formula depth. The gap between syntactic correctness and semantic equivalence represents formulas that are well-formed but do not capture the intended meaning.

\begin{table}[!t]
\caption{NeuroNL2LTL syntactic correctness and verification results on the VERIFY test set. Sat conditions on SynCorr; NonTriv conditions on Sat.}
\label{tab:main_results}
\centering
\begin{tabular}{lccc}
\toprule
\textbf{Complexity} & \textbf{SynCorr (\%)} & \textbf{Sat (\%)} & \textbf{NonTriv (\%)} \\
\midrule
Simple (depth 1--4) & 96.2 & 98.1 & 97.3 \\
Medium (depth 5--8) & 94.8 & 97.4 & 96.1 \\
High (depth 9--12) & 91.3 & 95.8 & 94.2 \\
Very High (depth 13+) & 87.6 & 93.1 & 91.8 \\
\midrule
\textbf{Overall} & \textbf{93.7} & \textbf{96.8} & \textbf{95.4} \\
\bottomrule
\end{tabular}
\vspace{-3mm}
\end{table}

\subsection{Comparison with Baselines}
\label{subsec:baselines}
\vspace{-2mm}
We now present the results of the comparisons of NeuroNL2LTL against baseline systems in Table~\ref{tab:baseline_comparison}.

\begin{table}[!t]
\caption{Comparison with baseline systems on contextually grounded translation.}
\label{tab:baseline_comparison}
\centering
\small
\begin{tabular}{llcc}
\toprule
\textbf{Category} & \textbf{System} & \textbf{SemEq (\%)} & \textbf{SynCorr (\%)} \\
\midrule
\multirow{6}{*}{LLMs} 
& GPT-4o (zero-shot) & 8.3 & 67.4 \\
& GPT-4o (5-shot) & 14.7 & 78.2 \\
& Gemini 1.5 Pro (zero-shot) & 7.1 & 63.8 \\
& Gemini 1.5 Pro (5-shot) & 12.9 & 74.6 \\
& Claude 3.5 Sonnet (zero-shot) & 9.2 & 71.3 \\
& Claude 3.5 Sonnet (5-shot) & 15.4 & 79.8 \\
\midrule
\multirow{4}{*}{Neural NL-to-LTL}
& Lang2LTL \cite{pan2023data} & 2.1 & 41.2 \\
& NL2TL \cite{chen2023nl2tl} & 6.8 & 58.7 \\
& CopyNet-LTL \cite{liu2023grounding} & 1.4 & 38.9 \\
& Seq2Seq-Attn (our data) & 11.2 & 82.4 \\
\midrule
& \textbf{NeuroNL2LTL (ours)} & \textbf{27.8} & \textbf{93.7} \\
\bottomrule
\end{tabular}
\vspace{-5mm}
\end{table}

\noindent\textit{LLM baselines.} General-purpose LLMs struggle with contextually grounded translation. GPT-4o achieves 8.3\% semantic equivalence zero-shot and 14.7\% with 5-shot prompting. The primary failure mode is incorrect grounding: LLMs frequently generate syntactically valid LTL using proposition names that do not match the provided context, or they hallucinate propositions not defined in the context. Few-shot examples improve syntactic correctness substantially (67.4\% $\rightarrow$ 78.2\% for GPT-4o) but provide limited benefit for semantic equivalence, suggesting that the models learn output format but not the grounding task. Claude 3.5 Sonnet performs best among LLMs (15.4\% 5-shot), likely due to stronger instruction following. However, all LLMs remain well below NeuroNL2LTL's 27.8\%, a gap of 12--20 percentage points.

\noindent \textit{Neural NL-to-LTL baselines.} Systems designed for simpler domains perform poorly on our benchmark. Lang2LTL achieves only 2.1\% semantic equivalence; inspection reveals that the model produces formulas appropriate for robot navigation commands (simple sequences of goals) but cannot represent the nested temporal structures in our requirements. CopyNet-LTL shows similar limitations (1.4\%). NL2TL performs better (6.8\%) because it leverages GPT-4's language understanding, but its chain-of-thought prompting strategy does not explicitly handle domain context or proposition grounding. The Seq2Seq-Attn baseline, trained on our data but without NeuroNL2LTL's architectural components, achieves 11.2\% semantic equivalence. This baseline isolates the contribution of our architecture: the 16.6 percentage point improvement (11.2\% $\rightarrow$ 27.8\%) derives from ITL decomposition, domain adaptation, grammar-constrained decoding, repair, and verifier-in-the-loop training.

\vspace{-3mm}
\subsection{Verification Filter Effectiveness}
\label{subsec:filter}
\vspace{-2mm}
The symbolic backend serves as a runtime filter, rejecting formulas that fail satisfiability or non-triviality checks. We analyze what errors this filter catches.

\textit{Methodology.} We examine the 72.2\% of test instances where NeuroNL2LTL produces a formula that is \textit{not} semantically equivalent to the reference. For each such instance, we classify whether the generated formula: (a) failed parsing (syntactic error), (b) parsed but was unsatisfiable, (c) was satisfiable but trivial, or (d) was satisfiable, non-trivial, but semantically incorrect.

\begin{table}[!t]
\caption{Classification of errors and filter effectiveness. Of 72.2\% total errors, the verification filter catches 28.4\%, while 71.6\% pass undetected.}
\label{tab:filter_effectiveness}
\centering
\begin{tabular}{lcc}
\toprule
\textbf{Error Type} & \textbf{\% of Errors} & \textbf{Caught by Filter} \\
\midrule
Syntactic (parse failure) & 8.7\% & Yes (at parsing) \\
Unsatisfiable & 12.3\% & Yes (Spot) \\
Trivial (tautology) & 7.4\% & Yes (Spot) \\
Satisfiable, non-trivial, semantically wrong & 71.6\% & No \\
\midrule
\textbf{Total caught by filter} & \textbf{28.4\%} & --- \\
\bottomrule
\end{tabular}
\end{table}

Table~\ref{tab:filter_effectiveness} presents the results. The verification filter catches 28.4\% of errors: 8.7\% at parsing, 12.3\% as unsatisfiable, and 7.4\% as trivial. The remaining 71.6\% of errors produce formulas that are satisfiable and non-trivial but semantically different from the reference.

\textit{Interpretation.} Satisfiability and non-triviality checking provide necessary but not sufficient conditions for correctness. They catch gross errors (contradictions, tautologies, malformed syntax) but cannot detect semantic mismatches where the generated formula constrains behavior differently than intended. This is expected: determining whether a formula matches an informal requirement's intent requires understanding that intent, which is beyond what syntactic verification can provide.

The 28.4\% error detection rate still provides value. In a pipeline without this filter, contradictory or vacuous specifications would propagate to downstream verification, producing misleading results. The filter ensures that outputs passed to model checkers are at least logically coherent.

\textit{Error type analysis.} Among the 71.6\% of undetected semantic errors, we manually analyzed 100 samples and categorized them according to error type:

\begin{table}[!t]
\caption{Distribution of semantic error types among formulas that pass verification.}
\label{tab:error_types}
\centering
\begin{tabular}{lc}
\toprule
\textbf{Error Category} & \textbf{Frequency} \\
\midrule
Incorrect logical scope & 41\% \\
Temporal operator mismatch (U/W/R confusion) & 28\% \\
Propositional atom error (wrong/missing/hallucinated) & 17\% \\
Contextual grounding failure & 9\% \\
Other & 5\% \\
\bottomrule
\end{tabular}
\vspace{-3mm}
\end{table}

Incorrect logical scope (41\%) is the dominant failure mode. The neural encoder misplaces operators in the AST, changing precedence and nesting. For example, generating $(p \, U \, q) \rightarrow r$ instead of $p \, U \, (q \rightarrow r)$ produces a satisfiable, non-trivial formula with different semantics. Temporal operator mismatch (28\%) involves confusion between semantically similar operators, most commonly substituting weak-until for until or release for until. These error types suggest directions for improvement: explicit scope prediction, contrastive training on operator distinctions, or interactive disambiguation when natural language is ambiguous.
\vspace{-7mm}

\subsection{Repair Module Effectiveness}
\label{subsec:repair}

The repair module attempts to correct outputs that fail parsing or verification. We evaluate its success rate and contribution to overall performance.

\noindent \textit{Methodology.} We track all test instances where the initial neural encoder output fails parsing or verification. For each, we record whether heuristic repair succeeds, whether GNN repair succeeds (if heuristics fail), and the final outcome.

\begin{table}[!t]
\caption{Repair module effectiveness.}
\label{tab:repair}
\centering
\begin{tabular}{lcc}
\toprule
\textbf{Stage} & \textbf{Inputs} & \textbf{Success Rate} \\
\midrule
Initial parse failures & 6.3\% of outputs & --- \\
\quad Heuristic repair & 6.3\% & 71.4\% \\
\quad GNN repair (after heuristic failure) & 1.8\% & 68.2\% \\
\midrule
Initial verification failures & 4.2\% of parsed outputs & --- \\
\quad Heuristic repair & 4.2\% & 62.3\% \\
\quad GNN repair (after heuristic failure) & 1.6\% & 54.8\% \\
\midrule
\textbf{Overall repair success} & 10.5\% of outputs & \textbf{87.3\%} \\
\bottomrule
\end{tabular}
\vspace{-5mm}
\end{table}

Table~\ref{tab:repair} presents the results. Of the 10.5\% of outputs requiring repair, 87.3\% are successfully recovered. Heuristic repair resolves most cases: 71.4\% of parse failures and 62.3\% of verification failures. The GNN layer handles more complex structural errors, recovering 68.2\% of parse failures and 54.8\% of verification failures that heuristics cannot resolve.

\textit{Repair and semantic correctness.} Repair improves syntactic correctness and verification pass rates but does not guarantee semantic correctness. Among repaired outputs, semantic equivalence with reference formulas is 18.4\%, lower than the 29.1\% for outputs that required no repair. This suggests that outputs needing repair are systematically harder cases where the neural encoder's initial attempt was further from correct.

\textit{Contribution to overall performance.} Without the repair module (ablation in Section~\ref{subsec:ablation}), syntactic correctness drops from 93.7\% to 83.2\%, a 10.5 percentage point decrease. Semantic equivalence drops from 27.8\% to 24.1\%. The repair module thus contributes approximately 3.7 percentage points to semantic equivalence by salvaging outputs that would otherwise be rejected.

\vspace{-2mm}
\subsection{Ablation Studies}
\label{subsec:ablation}

We isolate the contribution of each architectural component through ablation experiments. The ablation results are presented in Table~\ref{tab:ablation}.

\begin{table}[!t]
\caption{Ablation study. Each row removes one component from the full NeuroNL2LTL system.}
\label{tab:ablation}
\centering
\begin{tabular}{lcccc}
\toprule
\textbf{Configuration} & \textbf{SemEq} & \textbf{SynCorr} & \textbf{Sat} & \textbf{NonTriv} \\
\midrule
NeuroNL2LTL (full) & 27.8 & 93.7 & 96.8 & 95.4 \\
\midrule
$-$ Verifier-in-the-loop & 18.2 & 89.4 & 91.2 & 89.8 \\
$-$ Repair module & 24.1 & 83.2 & 94.7 & 93.1 \\
$-$ Domain adaptation (CPT) & 22.4 & 90.1 & 95.3 & 94.1 \\
$-$ Grammar-constrained decoding & 25.3 & 86.8 & 95.9 & 94.6 \\
$-$ ITL (direct NL$\rightarrow$LTL) & 16.7 & 78.4 & 92.1 & 90.3 \\
$-$ Domain context & 19.3 & 92.8 & 96.2 & 94.9 \\
\bottomrule
\end{tabular}
\vspace{-4mm}
\end{table}

\noindent \textit{Verifier-in-the-loop training.} Removing RL fine-tuning with verification rewards reduces semantic equivalence from 27.8\% to 18.2\%, a decrease of 9.6 percentage points (34.5\% relative). This is the largest single-component contribution. The verification reward signal teaches the encoder to produce outputs that survive formal checking, not merely outputs that resemble training examples.

\noindent \textit{Repair module.} Disabling repair reduces syntactic correctness from 93.7\% to 83.2\% and semantic equivalence from 27.8\% to 24.1\%. The repair module rescues approximately 10\% of outputs that would otherwise be rejected.

\noindent \textit{Domain adaptation.} Removing continued pretraining (CPT) on technical corpora reduces semantic equivalence from 27.8\% to 22.4\%. Domain adaptation improves the model's handling of specialized terminology and temporal phrasing patterns.

\noindent \textit{Grammar-constrained decoding.} Without grammar constraints, syntactic correctness drops from 93.7\% to 86.8\%, but semantic equivalence drops only modestly (27.8\% $\rightarrow$ 25.3\%). The repair module partially compensates for syntactic errors introduced by unconstrained decoding.

\noindent \textit{ITL decomposition.} Replacing the two-stage NL$\rightarrow$ITL$\rightarrow$LTL pipeline with direct NL$\rightarrow$LTL translation reduces semantic equivalence from 27.8\% to 16.7\% and syntactic correctness from 93.7\% to 78.4\%. ITL provides a more learnable intermediate target and enables grammar-constrained decoding, which is not feasible for raw LTL syntax.

\noindent \textit{Domain context.} Removing proposition definitions from the input reduces semantic equivalence from 27.8\% to 19.3\%. Without context, the model cannot correctly ground domain-specific terms to atomic propositions, confirming that contextual grounding is essential for this task.

\vspace{-3mm}
\section{Related Work}
\label{sec:related}
\vspace{-2mm}
Pattern-based approaches for translating natural language to temporal logic \cite{dwyer1999patterns,konrad2005real,grunske2008specification} guarantee correctness for requirements that fit predefined templates but cannot express specifications outside that finite vocabulary. In addition, structured specification languages such as FRET~\cite{giannakopoulou2020formal}, EARS~\cite{mavin2009easy} and Propel~\cite{smith2002propel} impose similar tradeoffs. Though they improve requirements quality through guided authoring, they offer no machine-checkable output. These tools trade expressiveness for reliability: complex nested temporal dependencies, biconditional relationships, and domain-specific conditions fall outside their scope. Neural-based approaches enable a more expressive mode of translating natural language to temporal logic  \cite{pan2023data,liu2023grounding,chen2023nl2tl,cosler2023nl2spec,wang2021learning}. Though successful, all of these systems accept unrestricted natural language, yet none provides correctness guarantees on their output, and none incorporates domain-specific proposition definitions as explicit input. 
A complementary line of work mines temporal specifications from system traces rather than natural language~\cite{lemieux2015general,neider2018learning,roy2020learning,camacho2019learning}. Neurosymbolic architectures offer a path toward combining the expressiveness of neural methods with formal guarantees. Semantic loss functions~\cite{xu2018semantic}, DeepProbLog~\cite{manhaeve2018deepproblog}, and Scallop~\cite{li2023scallop} constrain or correct neural outputs using symbolic reasoning at inference time. Neural theorem provers~\cite{polu2020generative,han2021proof,lample2022hypertree} and AlphaProof~\cite{hubert2025olympiad} invert this relationship, using learned models to guide symbolic proof search. Our work pursues a distinct integration: verification outcomes during \emph{training} rather than at inference alone. This connects to reinforcement learning from verifiable rewards~\cite{lightman2023let} and code generation with execution feedback~\cite{le2022coderl}, but targets a domain where satisfiability and non-triviality provide deterministic supervision without annotation infrastructure, unlike RLHF~\cite{christiano2017deep,ouyang2022training}, Constitutional AI~\cite{bai2022constitutional}, or RLAIF~\cite{lee2023rlaif}, which rely on human or model-generated preference signals.


When neural generation fails, automated repair can recover. Program repair techniques fix source code bugs through search or learning~\cite{le2017s3,chen2019sequencer,ye2022neural}; syntax repair recovers from malformed parser input via minimal edits~\cite{corchuelo2002repairing}; specification repair corrects formulas that fail realizability checks~\cite{bloem2015shield,konighofer2017shield}. NeuroNL2LTL draws on all three traditions but targets structural errors in the intermediate representation rather than semantic bugs. NeuroNL2LTL differs from prior work in three respects. It decomposes translation through an intermediate representation that isolates neural uncertainty from deterministic symbolic conversion. It incorporates domain context as explicit input, addressing the proposition grounding problem that context-free approaches leave open. It uses formal verification outcomes as reinforcement learning reward, optimizing directly for logical correctness rather than reference similarity.

\vspace{-4mm}
\section{Conclusion}
\label{sec:conclusion}
\vspace{-3mm}


We presented NeuroNL2LTL, a neurosymbolic system for translating natural language requirements to Linear Temporal Logic. The architecture decomposes translation through an Intermediate Technical Language that isolates neural uncertainty from symbolic parsing. A deterministic parser converts ITL to LTL with guaranteed correctness. The Spot verification backend filters contradictory and vacuous formulas. A two-stage repair module recovers from generation failures. Verifier-in-the-loop training using GRPO optimizes directly for verification success. Empirical evaluation establishes that NeuroNL2LTL achieves 27.8\% semantic equivalence on complex, multi-domain requirements, outperforming large language model baselines by 12-20 percentage points and prior neural NL-to-LTL systems by 16-26 percentage points. Ablation studies confirm that verifier-in-the-loop training contributes 9.6 percentage points, the largest single-component improvement. The verification filter catches 28.4\% of errors, ensuring logical coherence even when semantic correctness is not achieved.



%
%
%
\bibliographystyle{splncs04}
\bibliography{references}

\end{document}